\def\year{2020}\relax
\documentclass[letterpaper]{article} 
\usepackage{aaai20}  
\usepackage{times}  
\usepackage{helvet} 
\usepackage{courier}  
\usepackage[hyphens]{url}  
\usepackage{graphicx} 
\usepackage{graphicx}  
\usepackage{multirow}
\usepackage{amsmath}
\usepackage{amssymb}
\usepackage{bm}
\title{Bridging the Gap between Pre-Training and Fine-Tuning \\ for End-to-End Speech Translation}
\author{ \Large \textbf{Chengyi Wang \thanks{Works are done during internship at Microsoft},\textsuperscript{\rm 1} Yu Wu,\textsuperscript{\rm 2} Shujie Liu,\textsuperscript{\rm 2}  Zhenglu Yang,\textsuperscript{\rm 1} Ming Zhou\textsuperscript{\rm 2}}\\ 
\textsuperscript{\rm 1}Nankai University, Tianjin, China\\
\textsuperscript{\rm 2}Microsoft Research Aisa, Beijing, China\\
cywang@mail.nankai.edu.cn,  Wu.Yu@microsoft.com,  \\
shujliu@microsoft.com,  yangzl@nankai.edu.cn,  mingzhou@microsoft.com
\\ 
}
 
\begin{document}

\maketitle

\begin{abstract}
End-to-end speech translation, a hot topic in recent years, aims to translate a segment of audio into a specific language with an end-to-end model. Conventional approaches employ multi-task learning and pre-training methods for this task, but they suffer from the huge gap between pre-training and fine-tuning.   To address these issues, we propose a Tandem Connectionist Encoding Network (TCEN) which bridges the gap by reusing all subnets in fine-tuning, keeping the roles of subnets consistent, and pre-training the attention module. Furthermore, we propose two simple but effective methods to guarantee the speech encoder outputs and the MT encoder inputs are consistent in terms of semantic representation and sequence length. Experimental results show that our model leads to significant improvements in En-De and En-Fr translation irrespective of the backbones.
\end{abstract}

\section{Introduction}
Speech-to-Text translation (ST) is essential for a wide range of scenarios: for example in emergency calls, where agents have to respond emergent requests in a foreign language \cite{munro2010crowdsourced}; or in online courses, where audiences and speakers use different languages \cite{jan2018iwslt}. To tackle this problem, existing approaches can be categorized into cascaded method \cite{DBLP:conf/icassp/Ney99,DBLP:conf/acl/MaHXZLZZHLLWW19}, where a machine translation (MT) model translates outputs of an automatic speech recognition (ASR) system into target language, and end-to-end method \cite{duong2016attentional,DBLP:conf/interspeech/WeissCJWC17}, where a single model learns acoustic frames to target word sequence mappings in one step towards the final objective of interest. Although the cascaded model remains the dominant approach due to its better performance, the end-to-end method becomes more and more popular because it has lower latency by avoiding inferences with two models and  rectifies the error propagation in theory.

Since it is hard to obtain a large-scale ST dataset,  multi-task learning \cite{DBLP:conf/interspeech/WeissCJWC17,berard2018end} and pre-training techniques \cite{DBLP:conf/naacl/BansalKLLG19} have been applied to end-to-end ST model to leverage large-scale datasets of ASR and MT. A common practice is to pre-train two encoder-decoder models for ASR and MT respectively, and then initialize the ST model with the encoder of the ASR model and the decoder of the MT model. Subsequently, the ST model is optimized with the multi-task learning by weighing the losses of ASR, MT, and ST. This approach, however, causes a huge gap between pre-training and fine-tuning, which are summarized into three folds:

\begin{itemize}
\item  \textbf{Subnet Waste:}  The ST system just reuses the ASR encoder and the MT decoder, while discards other pre-trained subnets, such as the MT encoder. Consequently, valuable semantic information captured by the MT encoder cannot be inherited by the final ST system.
\item \textbf{Role Mismatch:} The speech encoder plays different roles in pre-training and fine-tuning. The encoder is a pure acoustic model in pre-training, while it has to extract semantic and linguistic features additionally in fine-tuning, which significantly increases the learning difficulty.

\item \textbf{Non-pre-trained Attention Module:} Previous work \cite{berard2018end} trains attention modules for ASR, MT and ST respectively, hence, the attention module of ST does not benefit from the pre-training.
\end{itemize}

\begin{figure*}
\centering
  \label{many-to-many}
  \includegraphics[width=0.85\textwidth]{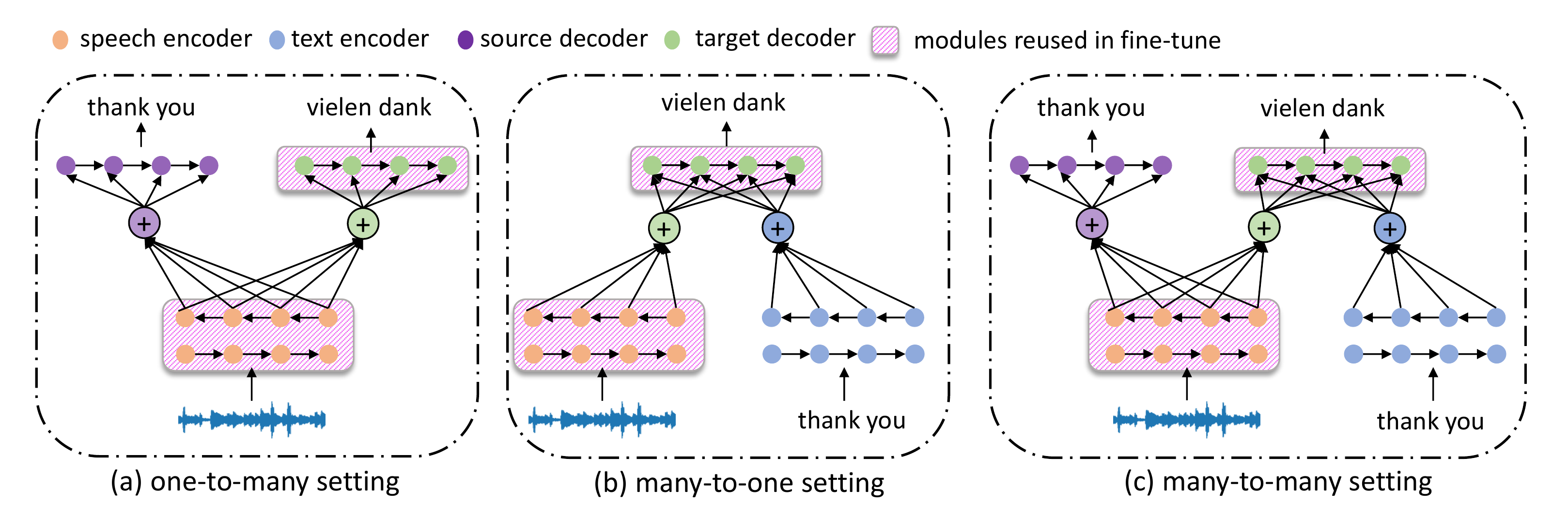}
  \caption{An illustration of multi-task learning for speech translation. Networks inherited from pre-trained models are labeled by  rectangles. }
\label{multitask}
\end{figure*}
To address these issues, we propose a Tandem Connectionist Encoding Network (TCEN), which is able to reuse all subnets in pre-training, keep the roles of subnets consistent, and pre-train the attention module.  Concretely, the TCEN consists of three components, a speech encoder, a text encoder, and a target text decoder. Different from the previous work that pre-trains an encoder-decoder based ASR model, we only pre-train an ASR encoder by optimizing the Connectionist Temporal Classification (CTC) \cite{DBLP:conf/icml/GravesFGS06} objective function. In this way, the additional decoder of ASR is not required while keeping the ability to read acoustic features into the source language space by the speech encoder. Besides, the text encoder and decoder can be pre-trained on a large MT dataset. After that, we employ common used multi-task learning method  to jointly learn ASR, MT and ST tasks.

Compared to prior works, the encoder of TCEN is a concatenation of an ASR encoder and an MT encoder and our model does not have an ASR decoder, so the \textbf{subnet waste issue} is solved. Furthermore, the two encoders work at tandem, disentangling acoustic feature extraction and linguistic feature extraction, ensuring the  \textbf{role consistency} between pre-training and fine-tuning. Moreover, we \textbf{reuse the pre-trained MT attention module} in ST, so we can leverage the alignment information learned in pre-training.

Since the text encoder consumes word embeddings of plausible texts in MT task but uses speech encoder outputs in ST task, another question is how one guarantees the speech encoder outputs are consistent with the word embeddings. We further modify our model  to achieve  \textbf{semantic consistency} and \textbf{length consistency}. Specifically, (1) the projection matrix at the CTC classification layer for ASR is shared with the word embedding matrix, ensuring that they are mapped to the same latent space, and (2)   the length of the speech encoder output is proportional to the length of the input frame, so it is much longer than a natural sentence. To bridge the length gap, source sentences in MT are lengthened by adding word repetitions and blank tokens to mimic the CTC output sequences.

We conduct comprehensive experiments on the IWSLT18 speech translation benchmark \cite{jan2018iwslt}, demonstrating the effectiveness of each component. Our model can lead to significant improvements for both LSTM and Transformer backbone.

Our contributions are three-folds: 1) we shed light on why previous ST models cannot sufficiently utilize the knowledge learned from the pre-training process; 2) we propose a new ST model, which alleviates shortcomings in existing methods; and 3) we empirically evaluate the proposed model on a large-scale public dataset.

 \begin{figure*}
\centering
  \includegraphics[width=0.8\textwidth]{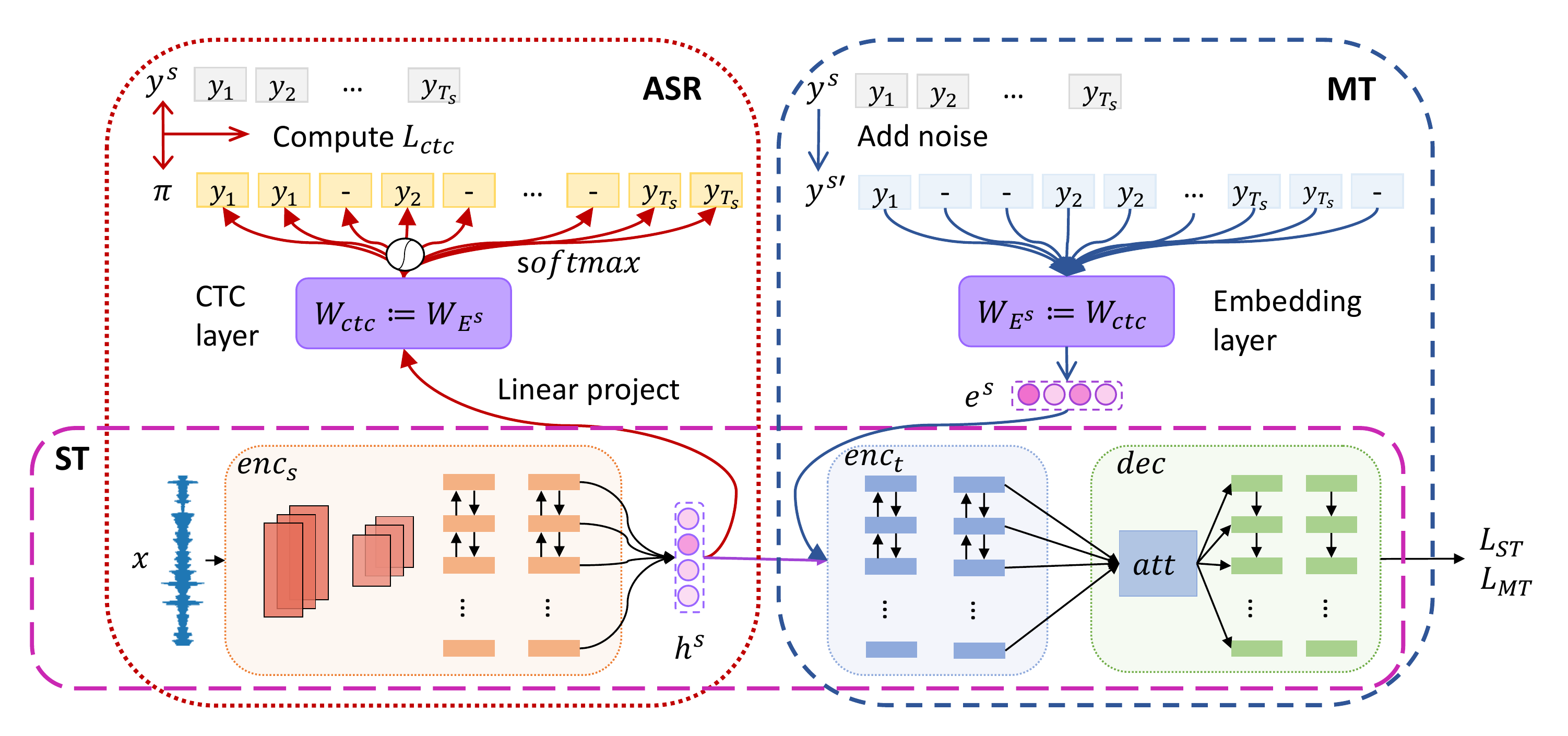}
  \caption{The architecture of our model. The linear projection matrix in ASR is shared with the word embedding matrix in MT. }
\label{model}
\end{figure*}

\section{Background}
End-to-end speech translation aims to translate a piece of audio into a target-language translation in one step. The raw speech signals are usually converted to sequences of acoustic features. Here, we define the speech feature sequence as $\bm{x} = (x_1, \cdots, x_{T_x})$.

The transcription and translation sequences are denoted as $\bm{y^{s}} = (y_1^{s}, \cdots, y_{T_s}^{s})$, and $\bm{y^{t}} = (y_1^{t}, \cdots, y_{T_t}^{t})$ repectively. Each symbol in $\bm{y^{s}}$ or $\bm{y^{t}}$ is an integer index of the symbol in a vocabulary $V_{src}$ or $V_{trg}$ respectively (e.g. $y^s_i=k, k\in [0, |V_{src}|-1]$). In this work, we suppose that an ASR dataset, an MT dataset, and a ST dataset are available, denoted as $\mathcal{A} = \{(\bm{x_i}, \bm{y^{s}_i})\}_{i=0}^I$, $\mathcal{M} =\{(\bm{y^{s}_j}, \bm{y^{t}_j})\}_{j=0}^J$ and  $ \mathcal{S} =\{ (\bm{x_l}, \bm{y^{t}_l})\}_{l=0}^L$ respectively.  Given a new piece of audio $\bm{x}$, our goal is to learn an end to end model to generate a translation sentence $\bm{y^{t}}$ without generating an intermediate result $\bm{y^{s}}$.

\subsection{Multi-Task Learning and Pre-training for ST}

To leverage large scale ASR and MT data, multi-task learning and pre-training techniques are widely employed to improve the ST system. As shown in Figure \ref{multitask}, there are three popular multi-task strategies for ST, including 1) one-to-many setting, in which a speech encoder is shared between ASR and ST tasks; 2) many-to-one setting in which a decoder is shared between MT and ST tasks; and 3) many-to-many setting where both the encoder and decoder are shared.

A many-to-many multi-task model contains two encoders as well as two decoders. It can be jointly trained on ASR, MT, and ST tasks. As the attention module is task-specific, three attentions are defined.

\section{Our method}
In this section, we first introduce the architecture of TCEN, which consists of two encoders connected in tandem, and one decoder with an attention module. Then we give the pre-training and fine-tuning strategy for TCEN. Finally, we propose our solutions for semantic and length inconsistency problems, which are caused by multi-task learning.

\subsection{Unified formulation for TCEN Architecture}
  Figure \ref{model} sketches the overall architecture of TCEN, including a speech encoder $enc_s$, a text encoder $enc_t$ and a decoder $dec$ with an attention module $att$. The $enc_s$ usually contains two modules: $\rm{EncPre}$ and $\rm{EncBody}$. During training, the $enc_s$ acts like an acoustic model which reads the input $\bm{x}$ to word or subword representations $\bm{h^s}$:
  \begin{small}
\begin{flalign}
    \bm{\tilde{x}}& = \text{EncPre}(\bm{x})\\
    \bm{h^s} & = \text{EncBody}(\bm{\tilde{x}})
\end{flalign}
\end{small}
  Then $enc_t$ learns high-level linguistic knowledge into hidden representations $\bm{h^t}$:
  \begin{small}
    \begin{equation}
      \bm{h^t} = \rm{enc_t}(\bm{h^s})
  \end{equation}
  \end{small}
Finally, the $dec$ defines a distribution probability over target words through attention mechanism:
\begin{small}
\begin{flalign}
    c_k & = \text{att}(z_{k-1}, \bm{h}^t) \\
    z_k & = \text{dec}(z_{k-1}, y_{k-1}^t, c_k) \\
    P(y_k^{t}|y_{<k}^t, x) &= \text{softmax}(W \cdot z_k)
\end{flalign}
\end{small}Here, $z_k$ is the the hidden state of the deocder at $k$ step and $c_k$ is a time-dependent context vector computed by the attention $att$.

The advantage of our architecture is that two encoders disentangle acoustic feature extraction and linguistic feature extraction, making sure that valuable knowledge learned from ASR and MT tasks can be effectively leveraged for ST training. However, there exists another problem:  In ST task, $enc_t$ accepts speech encoder output $\bm{h}^s$ as input. While in MT, $enc_t$ consumes the word embedding representation $\bm{e^s}$ derived from $\bm{y^s}$, where each element $e^s_i$ is computed by choosing the $y_i^s$-th vector from the source embedding matrix $W_{E^s}$. Since $\bm{h}^s$ and $\bm{e}^s$ belong to different latent space and have different lengths, there remain semantic and length inconsistency problems. We will provide our solutions in Section~\ref{subnet-consistency}. To verify the generalization of our framework, we test on LSTM based setting and Transformer \cite{DBLP:conf/nips/VaswaniSPUJGKP17} based setting.

\subsection{Training Procedure}
Following previous work, we split the training procedure to pre-training and fine-tuning stages. In pre-training stage, the speech encoder $enc_s$ is trained towards CTC objective using dataset $\mathcal{A}$, while the text encoder $enc_t$ and the decoder $dec$ are trained on MT dataset $\mathcal{M}$. In fine-tuning stage, we jointly train the model on ASR, MT, and ST tasks.

\begin{small}
\begin{table}
\scalebox{0.85}{
\begin{tabular}{l|l}
\hline
 Transcript $\bm{y^s}$    & we were not v @en @ge @ful at all\\ \hline
\multirow{2}{*}{CTC path $\bm{\pi_1}$} &  -(11) we we -(3) were -(3)  not  -(4) v @en\\
                  & @en @ge - @ful -(8) at at -(3) all -(10) \\ \hline
\multirow{2}{*}{CTC path $\bm{\pi_2}$} &  -(9) we -(3) were were -(4)  not not  -(3) v v @en\\
               &  @en @en @ge - @ful -(7) at -(3) all all -(10) \\ \hline
\end{tabular}}
\caption{An example of the comparison between the golden transcript and the predicted CTC paths given the corresponding speech. `-' denotes the blank token and the following number represents repeat times.}
\label{example}
\end{table}
\end{small}

\subsubsection{Pre-training} To sufficiently utilize the large dataset $\mathcal{A}$ and $\mathcal{M}$, the model is pre-trained on CTC-based ASR task and MT task in the pre-training stage.

For ASR task, in order to get rid of the requirement for decoder and enable the $enc_s$ to generate subword representation, we leverage connectionist temporal classification (CTC) \cite{DBLP:conf/icml/GravesFGS06} loss to train the speech encoder.

Given an input $\bm{x}$, $enc_s$ emits a sequence of hidden vectors $\bm{h^s}$, then a softmax classification layer predicts a CTC path $\bm{\pi}$, where $\pi_t \in V_{src} \cup$ \{`-'\} is the observing label at particular RNN step $t$, and `-' is the blank token representing no observed labels:
\begin{small}
\begin{equation}
    P(\bm{\pi}|\bm{x}) \approx \prod_{t=1}^{T}P(\pi_t|\bm{x}) = \prod_{t=1}^T \text{softmax}(W_{ctc} \cdot h^s_t)
\end{equation}
\end{small}where $W_{ctc} \in \mathbb{R}^{d \times (|V_{src}|+1)}$ is the weight matrix in the classification layer and $T$ is the total length of encoder RNN.

A legal CTC path $\bm{\pi}$ is a variation of the source transcription $\bm{y}^s$ by allowing occurrences of blank tokens and repetitions, as shown in Table \ref{example}. For each transcription $\bm{y}$, there exist many legal CTC paths in length $T$. The CTC objective trains the model to maximize the probability of observing the golden sequence $\bm{y}^s$, which is calculated by summing the probabilities of all possible legal paths:
\begin{small}
\begin{flalign}
     P(\bm{y}|\bm{x}) &= \sum_{\bm{\pi} \in \Phi_T(\bm{y})}P(\bm{\pi}|\bm{x}) \\
     \mathcal{L}_{CTC}(\bm{\theta}) &= -\sum_{(\bm{x}, \bm{y}^s) \in \mathcal{A}} \mathrm{log}P(\bm{y}^s|\bm{x}; \theta_{enc_s}, \theta_{W_{ctc}})
\end{flalign}
\end{small}
where $\Phi_T(y)$ is the set of all legal CTC paths for sequence $\bm{y}$ with length $T$. The loss can be easily computed using forward-backward algorithm. More details about CTC are provided in supplementary material.

 For MT task, we use the cross-entropy loss as the training objective. During training, $\bm{y^s}$ is converted to embedding vectors $\bm{e^s}$ through embedding layer $W_{E^s}$, then $enc_t$ consumes $\bm{e^s}$ and pass the output $\bm{h^t}$ to decoder. The objective function is defined as:
\begin{small}
\begin{equation}
    \mathcal{L}_{MT}(\bm{\theta})=-\sum_{(\bm{y}^s, \bm{y}^t) \in \mathcal{M}}\mathrm{log}\mathit{P}(\bm{y^{t}}|\bm{y^{s}};\theta_{enc_t}, \theta_{dec}, \theta_{W_{E^s}})
\end{equation}
\end{small}

\subsubsection{Fine-tune}
In fine-tune stage, we jointly update the model on ASR, MT, and ST tasks. The training for ASR and MT follows the same process as it was in pre-training stage.

For ST task, the $enc_s$ reads the input $\bm{x}$ and generates $\bm{h^s}$, then $enc_t$ learns high-level linguistic knowledge into $\bm{h^t}$. Finally, the $dec$ predicts the target sentence. The ST loss function is defined as:
\begin{small}
\begin{equation}
    \mathcal{L}_{ST}(\bm{\theta})=-\sum_{(\bm{x}, \bm{y}^t) \in \mathcal{S}}\mathrm{log}\mathit{P}(\bm{y^{t}}|\bm{x};\theta_{enc_s}, \theta_{enc_t}, \theta_{dec})
\end{equation}
\end{small}

Following the update strategy proposed by \citeauthor{luong2015multi} \shortcite{luong2015multi}, we allocate a different training ratio $\alpha_i$ for each task. When switching between tasks, we select randomly a new task $i$ with probability $\frac{\alpha_i}{\sum_{j}\alpha_{j}}$.

\subsection{Subnet-Consistency}\label{subnet-consistency}
Our model keeps role consistency between pre-training and fine-tuning by connecting two encoders for ST task.  However, this leads to some new problems:  1) The text encoder consumes $\bm{e^s}$ during MT training, while it accepts $\bm{h^s}$ during ST training. However, $\bm{e^s}$ and $\bm{h^s}$ may not follow the same distribution, resulting in the semantic inconsistency. 2) Besides, the length of $\bm{h^s}$ is not the same order of magnitude with the length of $\bm{e^s}$, resulting in the length inconsistency.

 In response to the above two challenges, we propose two countermeasures:  1) We share weights between CTC classification layer and source-end word embedding layer during training of ASR and MT, encouraging $\bm{e^s}$ and $\bm{h^s}$ in the same space.
2)We feed the text encoder source sentences in the format of CTC path, which are generated from a seq2seq model, making it more robust toward long inputs.

\subsubsection{Semantic Consistency}
As shown in Figure \ref{model}, during multi-task training, two different hidden features will be fed into the text encoder $enc_t$: the embedding representation $\bm{e}^s$ in MT task, and the $enc_s$ output $\bm{h^s}$ in ST task. Without any regularization, they may belong to different latent spaces. Due to the space gap, the $enc_t$ has to compromise between two tasks, limiting its performance on individual tasks.

To bridge the space gap, our idea is to pull $\bm{h^s}$ into the latent space where $\bm{e}^s$ belong. Specifically, we  share the weight $W_{ctc}$ in CTC classification layer with the source embedding weights $W_{E^s}$, which means $W_{ctc} = W_{E^s}$. In this way, when predicting the CTC path $\bm{\pi}$, the probability of observing the particular label $w_i \in V_{src}\cup $\{`-'\} at time step $t$, $p(\pi_t=w_i|\bm{x})$, is computed by normalizing the product of hidden vector $h_t^s$ and the $i$-th vector in $W_{E^s}$:
\begin{small}
\begin{equation}
    P(\pi_t=w_i|\bm{x}) = \frac{exp(W_{E^s}^{(i)} \cdot h^s_t)}{\sum_j^{|V_{src}|+1}exp(W_{E^s}^{(j)} \cdot h^s_t)}
\end{equation}
\end{small}
The loss function closes the distance between $h^s_t$ and golden embedding vector, encouraging $\bm{h}^s$ have the same distribution with $\bm{e}^s$.

\begin{table*}[t]
\centering
\scalebox{0.9}{
\begin{tabular}{c|c|c|c|c|c|c|c|c|c|c|c|c|c|c|c|c|c}
\hline
CTC path $\bm{\pi_1}$ & \multicolumn{17}{l}{-(11) we we -(3) were -(3)  not  -(4) v @en
               @en @ge - @ful -(8) at at -(3) all -(10)} \\ \hline
 $\bm{u}$  &  - & we & - &were &-& not& -& v& @en
                    &  @ge& - &@ful& -& at& - &all& -    \\ \hline
$\bm{l}$   &  11 &2 & 3  &1& 3 & 1& 4 & 1 & 2
               &  1 & 1 &  1 & 8 & 2 & 3 & 1& 10 \\ \hline
\end{tabular}}
\caption{The CTC path $\bm{\pi_1}$ and corresponding unique tokens $\bm{u}$ and repetition times $\bm{l}$, where $S(\bm{\pi}) = (\bm{u}, \bm{l})$.}
\label{example2}
\end{table*}

\begin{figure}
\centering
  \includegraphics[width=0.35\textwidth]{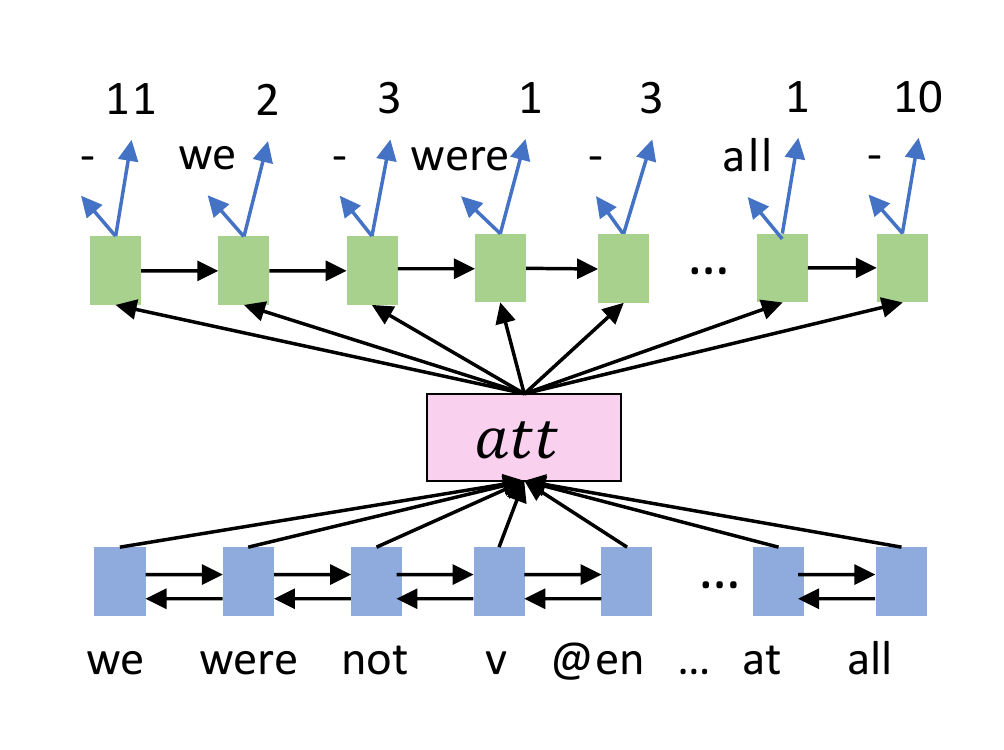}
  \caption{The architecture of seq2seq model. It predicts the next token and its number of repetition at the same time.}
\label{seq2seq}
\end{figure}
\subsubsection{Length Consistency}
Another existing problem  is length inconsistency. The length of the sequence $\bm{h^s}$ is proportional to the length of the input frame $\bm{x}$, which is much longer than the length of $\bm{e^s}$. To solve this problem, we train an RNN-based seq2seq model to transform normal source sentences to noisy sentences in CTC path format, and replace standard MT with denoising MT for multi-tasking.

Specifically, we first train a CTC ASR model based on dataset $\mathcal{A} = \{(\bm{x}_i, \bm{y}^s_i)\}_{i=0}^{I}$, and generate a CTC-path $\bm{\pi}_i$ for each audio $\bm{x}_i$ by greedy decoding. Then we define an operation $S(\cdot)$, which converts a CTC path $\bm{\pi}$  to a sequence of the unique tokens $\bm{u}$ and a sequence of repetition times for each token $\bm{l}$, denoted as $S(\bm{\pi}) = (\bm{u}, \bm{l})$. Notably, the operation is reversible, meaning that $S^{-1} (\bm{u}, \bm{l})=\bm{\pi}$.  We use the example $\bm{\pi_1}$ in Table \ref{example} and show the corresponding $\bm{u}$ and $\bm{l}$ in Table \ref{example2}.

Then we build a dataset $\mathcal{P} = \{(\bm{y^s}_i, \bm{u}_i, \bm{l}_i)\}_{i=0}^{I}$ by decoding all the audio pieces in $\mathcal{A}$ and transform the resulting path by the operation $S(\cdot)$.
After that, we train a seq2seq model, as shown in Figure \ref{seq2seq}, which takes $ \bm{y^s}_i$ as input and decodes  $\bm{u}_i, \bm{l}_i$ as outputs. With the seq2seq model, a noisy MT dataset  $\mathcal{M}'=\{(\bm{\pi}_l, \bm{y^t}_l)\}_{l=0}^{L}$ is obtained by converting every source sentence $\bm{y^s}_i \in \mathcal{M}$ to $\bm{\pi_i}$, where $\bm{\pi}_i = S^{-1}(\bm{u}_i, \bm{l}_i)$. We did not use the standard seq2seq model which takes $\bm{y^s}$ as input and generates $\bm{\pi}$ directly, since there are too many blank tokens `-' in $\bm{\pi}$ and the model tends to generate a long sequence with only blank tokens. During MT training, we randomly sample text pairs from $\mathcal{M}'$ and $\mathcal{M}$ according to a hyper-parameter $k$. After tuning on the validation set, about $30\%$ pairs are sampled from $\mathcal{M}'$. In this way, the $enc_t$ is more robust toward the longer inputs given by the $enc_s$.

\section{Experiments}

\subsection{Dataset}

\begin{small}
\begin{table*}\centering
\small
			\scalebox{1}{

\begin{tabular}{l|ccccc|ccccc}\hline
&   \multicolumn{5}{c|}{\textbf{Subword Level Decoder }}    &        \multicolumn{5}{c}{\textbf{Char Level Decoder }}        \\
\hline
                                      & tst2010 & tst2013 & tst2014 & tst2015 & Average& tst2010 & tst2013 & tst2014 & tst2015 & Average  \\ \hline
Vanilla                                 & 7.52    & 7.04    &  6.77   &  6.57    &  6.98  & 13.77  & 12.50    &  11.50   & 12.68   &  12.61    \\
$\,\,\,\,\,\,\,\,$+enc pretrain                    &   10.70      & 10.12   &    8.82     &    7.76     &   9.35     &  16.00       & 14.49   &   12.66      &    12.20     & 13.76   \\
$\,\,\,\,\,\,\,\,$+dec pretrain                    &   9.75      & 9.02    &   8.34      &    8.01     &  8.78     &  14.44  & 12.99    &   11.91& 12.87        &   13.05   \\
$\,\,\,\,\,\,\,\,$+enc dec pretrain           & 12.14   & 11.07   & 9.96    & 8.77    & 10.49    & 15.52   & 14.62   &  13.39   &  13.33   & 14.22 \\
pretrain+MTL       & 11.92   & 11.78   & 9.89    & 9.27    & 10.72  & 15.70   & 15.42   &  13.43 & 12.66   & 14.30  \\
Triangle+pretrain  & 9.89 & 9.91 & 7.48 & 7.22 & 8.63 & 11.35  & 10.73 & 9.43 & 9.47 &  10.25 \\

TCEN-LSTM                    & \bf{15.49}    &\bf{15.50}   & \bf{13.21}  & \bf{13.02}  & \bf{14.31}  & \bf{17.61}  & \bf{17.67}  & \bf{15.73}  & \bf{14.94}  & \bf{16.49}\\ \hline

\end{tabular}}
\caption{Results of LSTM-based models on ST TED. ``Average" denotes it averages the results of all test sets. We copy the numbers of vanilla model from \url{https://github.com/espnet/espnet/blob/master/egs/iwslt18/st1/RESULTS}. Since pre-training data is different, we run ESPnet code to obtain the numbers of pre-training and multi-task learning method, which are slightly higher than numbers in their report.}
\label{results}
\end{table*}
\end{small}
We conduct experiments on the Speech Translation TED (ST-TED) En-De corpus \cite{jan2018iwslt} and the augmented Librispeech En-Fr corpus \cite{DBLP:conf/lrec/KocabiyikogluBK18}.
\subsubsection{ST-TED En-De} The corpus contains 272 hours of English speech with 171k segments. Each example consists of raw English wave, English transcription, and aligned German translation.  Aside from ST-TED, we use TED-LIUM2 corpus \cite{rousseau2014enhancing} with 207h of speech data for ASR pre-training. For MT model, we use WMT2018 en-de data in pre-training stage and use sentence pairs in the ST-TED corpus as well as  WIT3\footnote{https://wit3.fbk.eu/mt.php?release=2017-01-trnted} in fine-tune stage. The pre-training data contains 41M sentence pairs and fine-tuning data contains 330k sentence paris in total.
We split 2k segments from the ST-TED corpus as dev set and tst2010, tst2013, tst2014, tst2015 are used as test sets.

\subsubsection{Librispeech En-Fr} This corpus is colleted by aligning e-books in French with English utterances, which contains 236 hours of speech in total. The English speech, English transcription, French text translations from alignment and Google Translate references are provided. Following previous work \cite{berard2018end}, we only use the 100 hours clean train set and double the training size by concatenating the aligned references with Google Translate references. We use the speech-transcription pairs and transcription-translation pairs for ASR and MT pre-training. No additional data is used. The dev set is used as validation set and we report results on the test set.

\subsubsection{Data preprocessing} Our acoustic features are 80-dimensional log-Mel filterbanks and 3-dimensional pitch features extracted with a step size of 10ms and window size of 25ms and extended with mean subtraction and variance normalization. The utterances with more than 3000 frames are discarded. All the sentences are in lower-case and the punctuation is removed. To increase the amount of training data, we perform speed perturbation on the raw signals with speed factors 0.9 and 1.1.

For the MT pre-training data, sentences longer than 80 words or shorter than 10 words are removed. Besides, we discard pairs whose length ratio between source and target sentence is smaller than 0.5 or larger than 2.0.
Word tokenization is performed using the Moses scripts\footnote{\url{https://github.com/moses-smt/mosesdecoder/blob/master/scripts/tokenizer/tokenizer.perl}} and all words are in lower-case.

For ST-TED experiments, we apply both subword-level decoding and character-level decoding. For the subword setting, both English and German vocabularies are generated using sentencepiece\footnote{https://github.com/google/sentencepiece} \cite{DBLP:conf/acl/Kudo18} with a fixed size of 5k tokens. For Librispeech En-Fr experiments, we only apply character-level decoding.

Since there are no human annotated alignments provided in ST-TED test sets, we segment each audio with the LIUM SpkDiarization tool \cite{meignier2010lium} and then perform MWER segmentation with RWTH toolkit \cite{DBLP:conf/iwslt/BenderZMN04}. Case-insensitive BLEU is used as evaluation metric.

\subsection{Experimental setups}
\subsubsection{Model architecture}

\noindent For LSTM based models, we follow the model structure in \citeauthor{inaguma2018speech} \shortcite{inaguma2018speech}. The $\rm{EncPre}$ corresponds to 2-layers of VGG-like max-pooling, resulting 4-fold downsampling of input feature. The $\rm{EncBody}$ is five bidirectional LSTM layers with cell size of 1024. The decoder is defined as two unidirectional LSTM layers with an additive attention. The decoder has the same dimension as the encoder RNNs.

 For Transformer based models, we use two-layer CNN with 256 channels, stride size 2 and kernel size 3 as $\rm{EncPre}$. The other modules are similar as in paper \citeauthor{DBLP:conf/icassp/DongXX18} \shortcite{DBLP:conf/icassp/DongXX18} ($e = 12, d=6, d_{model}=256, d_{ff} = 2048$ and $d_{head}=4$).

\subsubsection{Baselines} We compare our method with the following baselines.
\begin{itemize}
\item Vanilla ST baseline: The vanilla ST has only a speech encoder and a decoder, which is trained from scratch with only the speech-translation data.
\item Pre-training baselines: We conduct three pre-training baseline experiments: 1) encoder pre-training, 2) decoder pre-training, and 3) encoder-decoder pre-training. The pre-trained ASR model has the same architecture with vanilla ST model. The MT model has a $enc_t$ and $dec$ with the same architecture of which in TCEN.
\item  Pre-training + MTL: In this setting, we train a many-to-many multi-task model where the encoders and decoders are derived from pre-trained ASR and MT models.
\end{itemize}

\subsubsection{Implementation}

All our models are implemented based on ESPnet \cite{watanabe2018espnet}. For LSTM based models, we use a dropout of 0.3 for embedding and encoders. The model is trained using Adadelta  with initial learning rate of 1.0.

For Transformer based model, we use a dropout rate of 0.1 and a gradient clip of 5.0. Following \cite{}, we use Adam optimizer according to the learning rate schedule formula:
\begin{small}
\begin{equation*}
    \text{lrate} = k \cdot d_{model}^{-0.5} \cdot \min (n^{-0.5}, n \cdot warmup\_n^{-1.5})
\end{equation*}
\end{small}
We set $k=10$ and $warmup\_n = 25000$ in our experiments.
All the models are trained on 4 Tesla P40 GPU for a maximum of 20 epochs.

For training of TCEN, we set $\alpha_{asr}=0.2$ and $\alpha_{mt}=0.8$ in the pre-training stage, since the MT dataset is much larger than ASR dataset. For fine-tune, we use $\alpha_{st}=0.6, \alpha_{asr}=0.2$ and $\alpha_{mt}=0.2$. At inference time, we use a beam size of 10 and a length normalization weight of 0.2.

\subsection{Experimental Results}
 \subsubsection{Reults on ST TED}
 Table \ref{results} shows the LSTM-based results on four test sets as well as the average performance.  In this setting, we also re-implement the triangle multi-task strategy \cite{DBLP:conf/naacl/AnastasopoulosC18} as our baseline, denoted as `triangle+pretrain'. They concatenate a ST decoder to an ASR encoder-decoder model.

 From the table, we can see that our method significantly  outperforms the strong `pretrain+MTL' baseline by 3.6 and 2.2 BLEU scores respectively, indicating the proposed method is very effective.
 Besides, both pre-training and multi-task learning can improve translation quality. We observe a performance degradation in the `triangle+pretrain' baseline. Compared to our method, where the decoder receives higher-level knowledge extracted from text encoder, their ASR decoder can only provide lower word-level linguistic information. Besides, their model cannot utilize the large-scale MT data in all the training stages.
 \begin{table}\small
\scalebox{0.93}{
\begin{tabular}{l|l|l|l|l}
\hline
                     & tst2010 & tst2013 & tst2014 & tst2015 \\ \hline
cascaded             & 13.38  &  15.84  &  12.94  & 13.79   \\
cascaded+re-seg       & 17.12 & \bf{17.77}   & 14.94  &   \bf{15.01}      \\ \hline
our model  & \bf{17.61}  & 17.67  & \bf{15.73}  & 14.94  \\ \hline

\end{tabular}}
\caption{BLEU comparison of cascaded results and our best end-to-end results. re-seg denotes the ASR outputs are re-segmented before fed into the MT model.}
\label{cascade}
\end{table}

\begin{table}[t]
\centering
\begin{tabular}{l|c}
\hline
System                      & tst2013  \\ \hline
Vanilla               & -  \\
$\,\,\,\,$ +enc pretrain  &  13.41   \\
$\,\,\,\,$ +enc dec pretrain   &  14.46    \\
pretrain+MTL  & 14.98   \\
TCEN-Transformer & \bf{17.11} \\ \hline
\end{tabular}
\caption{BLEU of Transformer-based models on tst2013 set. `-': failed training.}
\label{Transformer}
\end{table}
Table \ref{cascade} shows the comparison between our best model with the cascaded systems, which combines the ASR model and MT model. In addition to a simple combination system, we also re-segment the ASR outputs before feeding to the MT system, denoted as `cascaded+re-seg'. Specifically, we train a seq2seq model \cite{DBLP:journals/corr/BahdanauCB14} on the MT dataset, where the source side is a no punctuation sentence and the target side is a natural sentence. After that, we use the seq2seq model  to add sentence boundaries and punctuation on ASR outputs.
It can be seen that our end-to-end model outperforms the simple cascaded model over 2 BLEU scores, and achieves a comparable performance with the `cascaded+re-seg' system.

 We list Transformer-based results on tst2013 in Table \ref{Transformer}. In this setting, we use character-level decoding strategy due to its better performance. Only in-domain MT data is used during pre-training. It can be seen that our TCEN framework works well on Transformer-based architecture and it outperforms the `pretrain+MTL' baseline by 2.1 BLEU scores.

 \subsubsection{Results on Librispeech}
\begin{small}
\begin{table}
\centering
\begin{tabular}{c|l|c}
\hline
\multicolumn{2}{c|}{Model}      & BLEU  \\ \hline
\multicolumn{1}{c|}{MT}   & \citeauthor{berard2018end}\shortcite{berard2018end}   & 19.2  \\
                          & ESPnet*   & 18.3  \\ \hline
Cascaded               & \citeauthor{berard2018end}\shortcite{berard2018end}   & 14.6  \\
                     & ESPnet*    & 15.8  \\ \hline
E2E                & \citeauthor{berard2018end}\shortcite{berard2018end}  & 12.9  \\
ST                     & \,\,\,\,+Pretrain+MTL  & 13.4  \\
& \citeauthor{DBLP:journals/corr/abs-1904-08075}\shortcite{DBLP:journals/corr/abs-1904-08075} & 17.02 \\
                      & ESPnet*  & 15.71 \\
                      & \,\,\,\,+enc pretrain     & 16.30 \\
                      & \,\,\,\,+enc dec pretrain      & 16.78 \\
                      &TCEN-LSTM & \bf{17.05}  \\ \hline
\end{tabular}
\caption{BLEU results of LSTM-based models on Librispeech En-Fr. *: The ESPnet baseline results are copied from \url{https://github.com/espnet/espnet/blob/master/egs/libri_trans}  }
\label{libirspeech}
\end{table}
\end{small}

For this dataset, we only perform LSTM-based experiments and report results in Table \ref{libirspeech}. Even without utilizing large-scale ASR data or MT data, our method can outperform the pre-training baselines and achieve the same performance with \citeauthor{DBLP:journals/corr/abs-1904-08779} \shortcite{DBLP:journals/corr/abs-1904-08779}, which uses a MT model as a teacher model to guide the ST model.
\subsection{Discussion}

\begin{table}[t]\small
\scalebox{1}{
\begin{tabular}{l|l|l|l|l}
\hline
System                     & tst2010 & tst2013 & tst2014 & tst2015 \\ \hline
TCEN              & 15.49 & 15.50  & 13.21  &  13.02 \\
$\,\,\,\,$ -MT noise  &  15.01  & 14.95   & 13.34 &  12.80 \\
$\,\,\,\,$ -weight sharing   &  13.51    & 14.02  & 12.25    & 11.66 \\
$\,\,\,\,$ -pretrain  & 8.98  & 8.42  &  7.94 & 8.08 \\ \hline
\end{tabular}}
\caption{Ablation study for subword-level experiments.}
\label{ablation}
\end{table}

\subsubsection{Ablation Study}
To better understand the contribution of each component, we perform an ablation study on subword-level experiments for ST TED corpus. The results are shown in Table \ref{ablation}. In `-MT noise' setting, we do not add noise to source sentences for MT. In `-weight sharing' setting, we use different parameters in CTC classification layer and source embedding layer. These two experiments prove that both weight sharing and using noisy MT input benefit to the final translation quality. Performance degrades more in `-weight sharing', indicating the semantic consistency contributes more to our model.

In the `-pretrain' experiment, we remove the pre-training stage and directly update the model on three tasks, leading to a dramatic decrease on BLEU score, indicating the pre-training is an indispensable step for end-to-end ST.

\begin{figure}
\centering
  \includegraphics[width=0.4\textwidth]{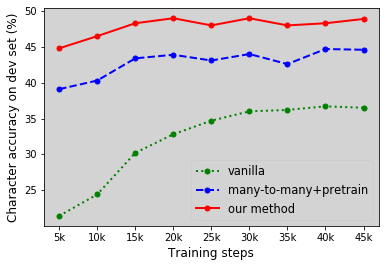}
  \caption{Model learning curves in fine-tuning. }
\label{learning curve}
\end{figure}
\subsubsection{Learning Curve} It is interesting to investigate why our method is superior to baselines.  We find that TCEN achieves a higher final result owing to a better start-point in fine-tuning.
Figure \ref{learning curve} provides learning curves of subword accuracy on validation set. The x-axis denotes the fine-tuning training steps. The vanilla model starts at a low accuracy, because its networks are not pre-trained on the ASR and MT data. The trends of our model and `many-to-many+pretrain' are similar, but our model outperforms it about five points in the whole fine-tuning process. It indicates that the gain comes from bridging the gap between pre-training and fine-tuning rather than a better fine-tuning process.

\section{Related Work}
Early works conduct ST in a pipeline manner \cite{DBLP:conf/icassp/Ney99,DBLP:conf/interspeech/MatusovKN05}, where the ASR output are fed into an MT system  to generate target sentences. HMM \cite{juang1991hidden}, DenseNet \cite{huang2017densely}, TDNN \cite{peddinti2015time} are commonly used ASR systems, while RNN with attention \cite{DBLP:journals/corr/BahdanauCB14} and Transformer \cite{DBLP:conf/nips/VaswaniSPUJGKP17} are top choices for MT. 

To avoid error propagation and high latency issues, recent works propose translating the acoustic speech into text in target language without yielding the source transcription  \cite{duong2016attentional,DBLP:journals/corr/BerardPSB16}. Since ST data is scarce, pre-training~\cite{DBLP:conf/naacl/BansalKLLG19}, multi-task learning \cite{duong2016attentional,berard2018end}, curriculum learning \cite{DBLP:journals/corr/abs-1802-06003}, attention-passing \cite{DBLP:journals/tacl/SperberNNW19}, and knowledge distillation \cite{DBLP:journals/corr/abs-1904-08075,jia2019leveraging} strategies have been explored to utilize ASR data and MT data.
Specifically, \citeauthor{DBLP:conf/interspeech/WeissCJWC17} \shortcite{DBLP:conf/interspeech/WeissCJWC17} show improvements of performance by training the ST model jointly with the ASR and the MT model. \citeauthor{berard2018end} \shortcite{berard2018end} observe faster convergence and better results due to pre-training and multi-task learning on a larger dataset. \citeauthor{DBLP:conf/naacl/BansalKLLG19} \shortcite{DBLP:conf/naacl/BansalKLLG19} show that pre-training a speech encoder on one language can improve ST quality on a different source language. All of them follow the traditional multi-task training strategies. \citeauthor{DBLP:journals/corr/abs-1802-06003} \shortcite{DBLP:journals/corr/abs-1802-06003} propose to use curriculum learning to improve ST performance on syntactically distant language pairs. To effectively leverage transcriptions in ST data, \citeauthor{DBLP:conf/naacl/AnastasopoulosC18} \shortcite{DBLP:conf/naacl/AnastasopoulosC18} augment the multi-task model where the target decoder receives information from the source decoder and they show improvements on low-resource speech translation. \citeauthor{DBLP:conf/icassp/JiaJMWCCALW19} \shortcite{DBLP:conf/icassp/JiaJMWCCALW19} use pre-trained MT and text-to-speech (TTS) synthesis models to convert weakly supervised data into ST pairs and demonstrate that an end-to-end MT model can be trained using only synthesised data.

\section{Conclusion}
This paper has investigated the end-to-end method for ST. We propose a method to reuse every sub-net and keep the role of sub-net consistent between pre-training and fine-tuning, alleviating the gap between pre-training and fine-tuning in previous methods. Empirical studies have demonstrated that our model significantly outperforms baselines.
\section{Acknowledgements}
This work was supported in part by the National Natural Science Foundation of China under Grant No.U1636116, 11431006.

\bibliography{aaai2020}
\bibliographystyle{aaai}
\end{document}